\documentclass[runningheads]{llncs}
\usepackage{graphicx}
\usepackage{amsmath}
\usepackage{subfigure}
\begin{document}
\title{SegViz: A federated-learning based framework for multi-organ segmentation on heterogeneous data sets with partial annotations}
\author{Adway Kanhere\inst{1,2} \and
Pranav Kulkarni\inst{2} \and
Paul H. Yi\inst{2} \and
Vishwa S. Parekh\inst{2}}

\institute{Department of Biomedical Engineering, Johns Hopkins University, Baltimore, MD \\
\email{akanher1@jhu.edu} \and
University of Maryland Medical Intelligent Imaging (UM2ii) Center, Baltimore, MD \\
\email{\{akanhere,pkulkarni,pyi,vparekh\}@som.umaryland.edu}}
\maketitle 
\begin{abstract}
Segmentation is one of the most primary tasks in deep learning for medical imaging, owing to its multiple downstream clinical applications. However, generating manual annotations for medical images is time-consuming, requires high skill, and is an expensive effort, especially for 3D images. One potential solution is to aggregate knowledge from partially annotated datasets from multiple groups to collaboratively train global models using Federated Learning. To this end, we propose SegViz, a federated learning-based framework to train a segmentation model from distributed non-i.i.d datasets with partial annotations. The performance of SegViz was compared against training individual models separately on each dataset as well as centrally aggregating all the datasets in one place and training a single model. The SegViz framework using FedBN as the aggregation strategy demonstrated excellent performance on the external BTCV set with dice scores of 0.93, 0.83, 0.55, and 0.75 for segmentation of liver, spleen, pancreas, and kidneys, respectively, significantly ($p<0.05$) better (except spleen) than the dice scores of 0.87, 0.83, 0.42, and 0.48 for the baseline models. In contrast, the central aggregation model significantly ($p<0.05$) performed poorly on the test dataset with dice scores of 0.65, 0, 0.55, and 0.68. Our results demonstrate the potential of the SegViz framework to train multi-task models from distributed datasets with partial labels. All our implementations are open-source and available at https://anonymous.4open.science/r/SegViz-B746
\keywords{Federated Learning  \and Partial annotations \and Segmentation.}
\end{abstract}
\section{Introduction}
% \subsection{A Subsection Sample}
Medical image segmentation is one of the most fundamental tasks in automated medical image analysis as it forms the basis for many downstream applications, including diagnosis, prognosis and treatment planning, image reconstruction, and treatment response assessment \cite{chen2021deep}. As a result, many large-scale datasets have been curated and released for the segmentation of different organ types and tumor structures \cite{antonelli2022medical}. However, each of these datasets has been curated for a specific use case and therefore, focuses on segmenting only a particular organ or tumor subset in the body. Consequently, developing and deploying algorithms for each use case would potentially result in hundreds of models, thereby limiting their clinical utility – imagine deploying a different algorithm for every type of cancer, injury, and other diseases. Considering the above limitations, the situation is further amplified by the time-consuming and expensive manual annotations required to build large-scale fully annotated multi-organ datasets. 

\begin{figure}[htp]
    \centering
    \includegraphics[width=8cm]{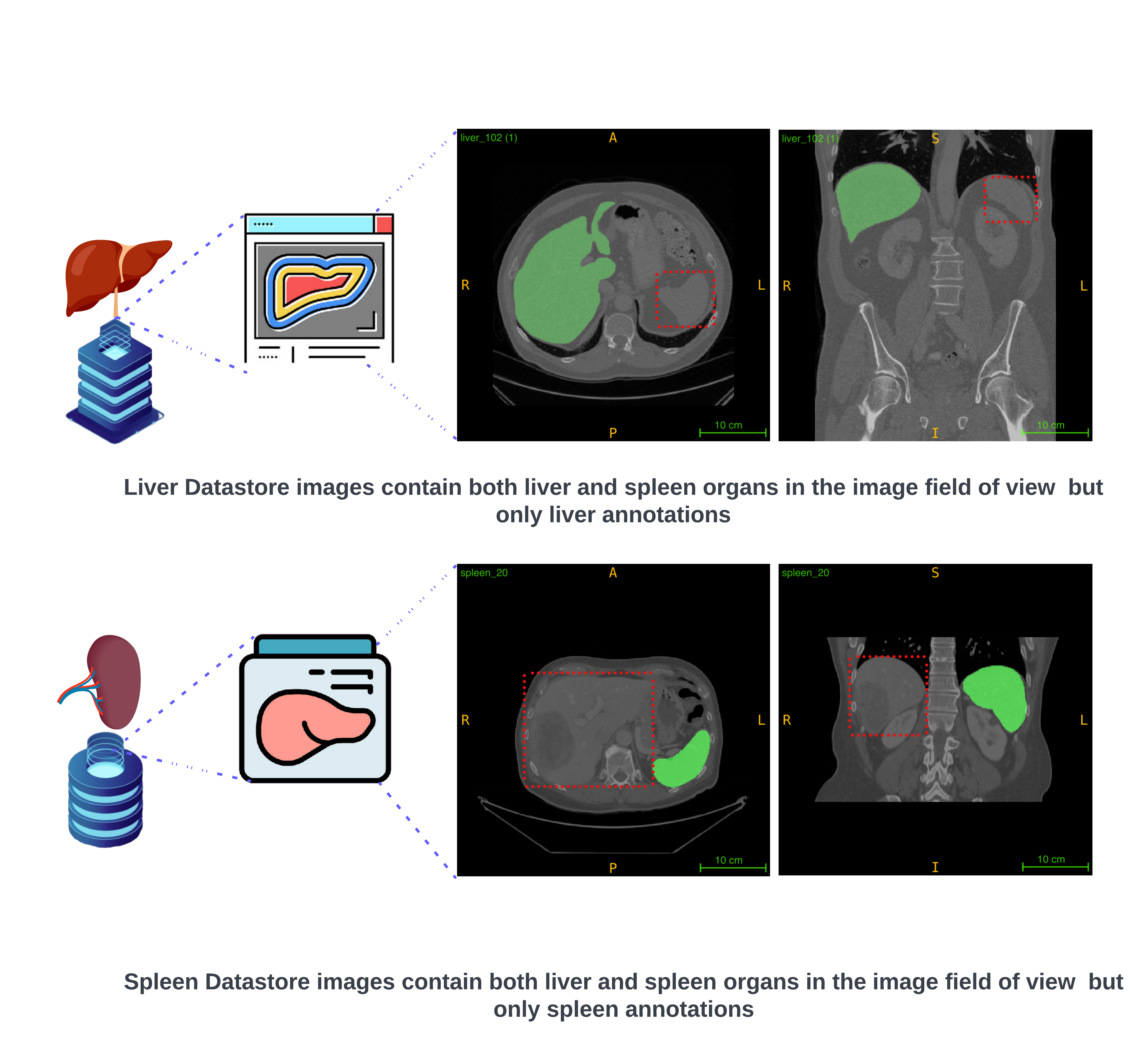}
    \caption{Illustration of an example federated learning setup with nodes containing datasets with a similar field of view but different and incomplete annotations.}
    \label{fig:introfig}
\end{figure}

The challenge of training several individual models separately and the need for large-scale multi-organ datasets can be addressed by training multi-task segmentation models from distributed datasets using collaborative learning. Federated learning (FL) has gained importance in recent years for solving this challenge by collaboratively training one global model from several local models without data sharing. However, the capability of FL in aggregating knowledge from datasets curated at different imaging centers is challenging as each imaging center may focus on related but different tasks; suppose one center is training a liver segmentation model while another center is training a spleen segmentation model from CT scans. These two datasets would contain images with a similar field of view but different, incomplete annotations, as illustrated in \ref{fig:introfig}. Such a situation, where one dataset has only a few organs annotated while another dataset contains no overlapping annotations with the first one is very common in medical imaging.

In this paper, we propose SegViz, a federated learning (FL) based framework for aggregating knowledge from heterogeneous, distributed medical imaging datasets with distinct and partial annotations into a single ‘global‘ model. We evaluated the SegViz framework for the task of segmenting four organs - liver, spleen, pancreas, and kidneys on CT scans using distributed nodes each containing one local dataset. We compare the performance of SegViz trained global models to models trained individually on each dataset as well as a model trained by centrally aggregating all the datasets. 

\section{Related Work}
Generating manual annotations for medical images is time-consuming, requires high skill, and is an expensive effort, especially for 3D images \cite{tajbakhsh2020embracing}. One potential solution is to curate datasets with partial annotations, wherein only a subset of structures is annotated for each image or volume. Furthermore, knowledge from similar partially annotated datasets from multiple groups can be aggregated to collaboratively train global models using Federated Learning \cite{chowdhury2022review}. Knowledge aggregation would not only save time but also allow different groups to benefit from each other's annotations without explicitly sharing them. Consequently, different techniques have been proposed in the literature for aggregating knowledge from heterogeneous datasets with partial, incomplete labels \cite{parekh2021cross,shen2021multi,boutillon2022generalizable,shen2022joint}.

There has been considerable research in the past on developing multi-task segmentation models using partial labels. The works of \cite{yu2019crossbar,isensee2019automated} show how to create subsets of the partially labeled datasets to create fully labeled subsets. However, this strategy requires very heavy computational resources. Another approach as described by \cite{huang2020multi,chen2019med3d} is to design a multi-task head with a common encoder and task-specific decoders that are trained separately. Similarly, the work of \cite{zhang2021dodnet,fang2020multi} has shown promise in developing multi-task segmentation models using multi-scale feature abstraction. However, these approachs require all the data to be hosted locally and is not realistic in a medical scenario not only because of privacy and data sharing restrictions but also because it is impossible to anticipate in advance how many distinct activities should the model be trained for.

In \cite{boutillon2022generalizable}, the authors developed a multi-task multi-domain deep segmentation model for the segmentation of pediatric imaging datasets with excellent performance. However, the proposed technique was developed and evaluated for different anatomical regions in the body with no overlapping field of view or incomplete annotations. Similarly, the cross-domain medical image segmentation technique developed in \cite{parekh2021cross} was focused on segmentation of the same anatomical structure and the proposed technique was not developed to tackle incomplete annotations

The work of \cite{xu2022federated} introduced a real FL setup for segmentation using partial labels where client nodes were trained on specific sub-networks for their specific tasks using a shared decoder. However, this method is not scalable and again, needs knowledge of all the tasks to be trained. It was for the first time in the work of \cite{shen2022joint} that knowledge aggregation was introduced using a single network in a federated manner. The global federated learning framework developed in their work, however, failed to accurately segment different anatomical structures on the external test set. For optimal performance, the authors used an ensemble of multiple local federated learning models, making it computationally expensive and practically challenging. 

Therefore, we developed SegViz to address the shortcomings of current techniques in efficiently aggregating knowledge from heterogeneous datasets with partial annotations. Our method does not rely on any heavy model or specific feature engineering methods and utilizes the intrinsic similarities between the different imaging datasets to learn a general representation across multiple tasks. Moreover, it does not require knowledge of the all the tasks in the participating datasets and is able to tackle domain shift between these datasets.

\section{Methods}
\begin{figure}[htp]
\centering
\includegraphics[width=10cm]{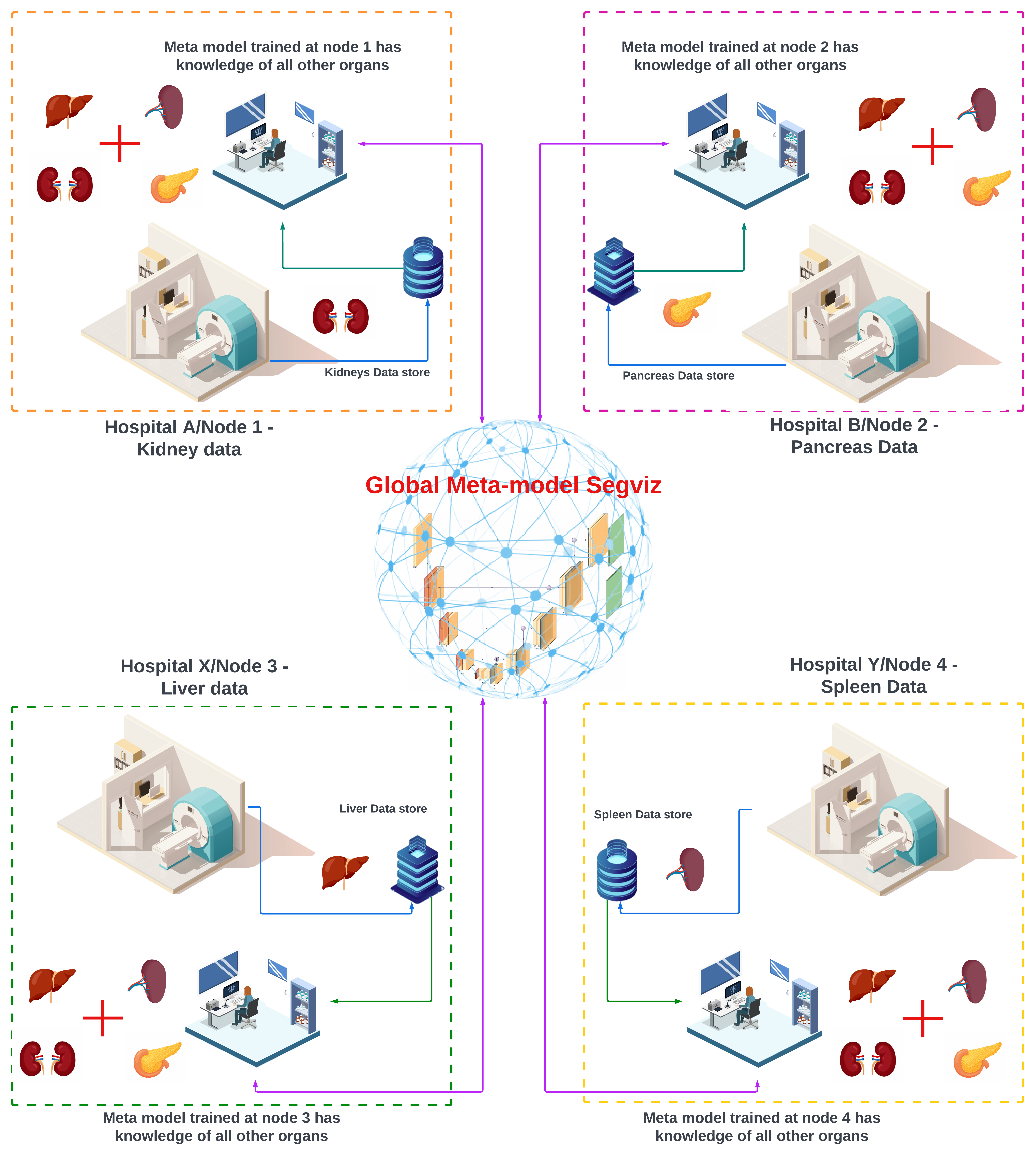}
\caption{Illustration of the proposed SegViz framework: Client nodes update the global meta-model where knowledge aggregation occurs after every 10 iterations of the local model. The weights of the global model are then shared with the client models allowing both nodes to share knowledge without sharing data.}
\label{fig:subim2}
\end{figure}

We developed SegViz as a multi-task federated learning framework to learn a diverse set of tasks from distributed nodes with incomplete annotations, as illustrated in Figure \ref{fig:subim2}. The global SegViz model is initialized at the server with two distinct blocks - a representation block and a task block. The goal of the representation block is to learn a generalized representation of the underlying dataset while the goal of the task block is to learn individual tasks distributed across different nodes. Every client is initialized with a subset of the SegViz model, comprising the representation block and a subset of the task block representing the client's tasks. During training, the weights of the representation block are always aggregated by the server and redistributed back to the client nodes. On the other hand, the weights of the task block are directly copied from the corresponding client nodes containing the corresponding task, thereby preserving the task-related information for each node in their task block. 
\subsection{SegViz model architecture}
The backbone of the SegViz model architecture was constructed using a modified version of the multi-head 3D-UNet \cite{cciccek20163d} configuration for all our experiments. Each U-Net has 5 layers with down/up-sampling at each layer by a factor of 2. Unlike how U-Net implementations typically operate, these down or up-sampling operations happen at the beginning of each block instead of at the end. The U-Net also contains 2 convolutional residual units at the layers and uses Batch Normalization at each layer. The task block comprised a multi-head architecture with each head consisting of two layers, including the final classification layer. The SegViz model was implemented using the MONAI \cite{cardoso2022monai} framework and the pre-processing and training were done using Pytorch. The SegViz model architecture has been illustrated in the supplementary material. 

During training, all weights are initialized using LeCun initialization. The batch size was set to 2 and the learning rate was initially set to 1e-4 with the Adam optimizer and CosineAnnealingLR \cite{loshchilov2016sgdr} as the scheduler. The Dice Loss was used as the loss function. The average Dice Score was chosen as the final evaluation metric. Each model was trained for a total of 500 epochs.

\subsection{Data}
The SegViz framework was evaluated using four publicly available datasets from the Medical Segmentation Decathalon (MSD) challenge \cite{antonelli2022medical}. The Spleen MSD dataset consists of 61 3D Computed tomographies (CT) volumes with spleen annotations out of which only the 41 training set volumes were used. The Liver MSD dataset consists of 201 3D CT volumes with liver and liver tumor annotations out of which only 131 training set volumes were considered. Similarly, the Pancreas MSD dataset consists of 420 3D CT volumes of which only 282 training volumes were used. Lastly, from the 2019 Kidney Tumor Segmentation Challenge dataset \cite{heller2019kits19}, we used 210 3D CT volumes from the training dataset. For this study, all tumor annotations were discarded and only organ annotations were used. The training and internal validation splits were considered from the overall training data in an 80:20 split. 
We considered all 30 training image volumes from the Beyond the Cranial Vault (BTCV) dataset \cite{landman2015miccai} as an external test set for all our experiments. 

During pre-processing, all the image volumes were first resized to 256 × 256 × 128, and the intensity values normalized between 0 and 1. All the volumes were resampled to a constant spacing of (1.5, 1.5, 2.0). We extract random foreground patches of size 128 ×128 × 32 from each volume such that the center voxel of each patch belonged to either the foreground or background class. 

\section{Experiments}
\subsection{Individual baseline implementation}
For every task, we trained a single U-Net model based on the Segviz model architecture on the training dataset after the 80:20 split. Hence we had a single model trained on the training dataset for the liver, spleen, pancreas, and kidneys. 
\subsection{Central aggregation implementation}
As a lower bound for a multi-task segmentation setup, we combine all four datasets together to create a central repository of all the data. We consider this a lower bound because naive aggregation of the data in the case of partial annotations would lead to suboptimal performance compared to an individual model trained for each dataset separately. We setup our centrally aggregated model using the same steps as our baseline implementation.

\subsection{SegViz implementation}
\textbf{FedAvg: }We use the popular FedAvg algorithm to construct an FL setup where each client node in the setup represents an isolated group having one of the datasets. The same UNet configuration with the Segviz architecture was used at each client. Apart from the same pre-processing steps as the baseline implementation, we also added random affine transformations such as rotation and scaling. While training the local models, after every 10 epochs, following the FedAvg algorithm, the global model gets all but the last two convolutional layers’ weights and averages them. The updated weights are then shared back to all the local models. 

\textbf{FedBN: } We investigate the popular FedBN algorithm in a similar setup as our FedAvg implementation. Making sure that our global model is generalizable to non-i.i.d data is especially important in medical imaging as data from different centers is obtained using different scanners/protocols. FedBN has shown to be successful compared to other FL algorithms such as FedAvg and FedProx in creating a global model that is generalizable well to non-i.i.d data and it does so by not aggregating the batch norm layers during knowledge transfer. 

\textbf{Local fine-tuning}
In \cite{jiang2019improving,liang2020think}, the authors demonstrated the need for fine-tuning in FL models in order to reduce the effect of catastrophic forgetting and stabilize personalized performance. We also finetuned our FedAvg and FedBN models (keeping the representation block frozen) on the local datasets to improve task-specific performance of each task block while keeping the same representation block. 

\begin{table}[!htb]
    \centering
    \caption{Mean Dice score performance of all the experiments on the in-federation validation dataset. The standard deviation values are in parentheses. After a paired t-test, the entries underlined are significant against baseline, in bold against central agg, in italics against our best model (FedBN + FT)}
    \includegraphics[width=\linewidth]{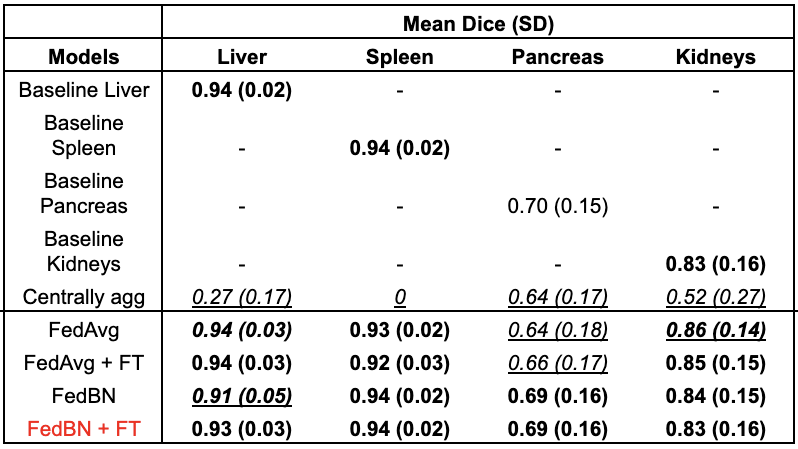}
    \label{tab:valid_metrics}
\end{table}

\section{Results}
\begin{figure}[htp]
\centering
\includegraphics[width=\linewidth, height = 6cm]{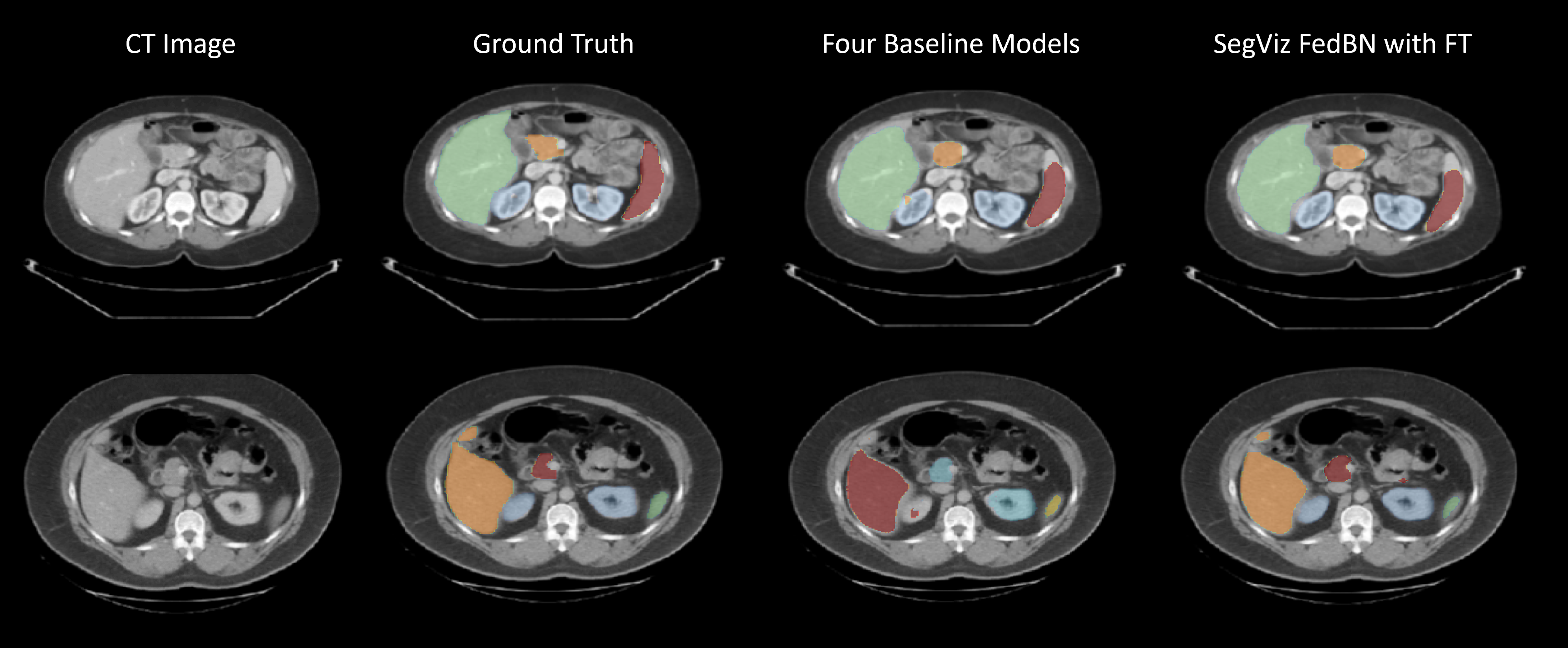}
\caption{A comparison of the ground truth segmentation masks with the masks generated by the baseline and SegViz models.}
\label{fig:overlays}
\end{figure}

As shown in Figure \ref{fig:overlays}, the FedBN model with fine-tuning performs the best on the in-federation internal validation set as well the out-of-federation BTCV test set. The SegViz framework using FedBN with fine-tuning segmented the BTCV test set with dice scores of 0.93, 0.83, 0.55, and 0.75 for segmentation of liver, spleen, pancreas, and kidneys, respectively, significantly ($p<0.05$) better (except spleen) than the dice scores of 0.87, 0.83, 0.42, and 0.48 for the baseline models. In contrast, the central aggregation model performed significantly ($p<0.05$) poorly on the test dataset with dice scores of 0.65, 0, 0.55, and 0.68. We note that the model trained on the centrally aggregated data did not generalize to the spleen label due to the overall model becoming biased toward the liver and pancreas labels, which contain more samples per label. We have included the statistical t-test results between the baseline and the best-performing models in Table \ref{tab:valid_metrics} and Table \ref{tab:btcv_metrics}. 

\begin{table}[!htb]
    \centering
    \caption{Mean Dice score performance of all the experiments on the out-of-federation BTCV dataset. The standard deviation values are in parentheses. The same conventions as Table 1 are followed}
    \includegraphics[width=\linewidth]{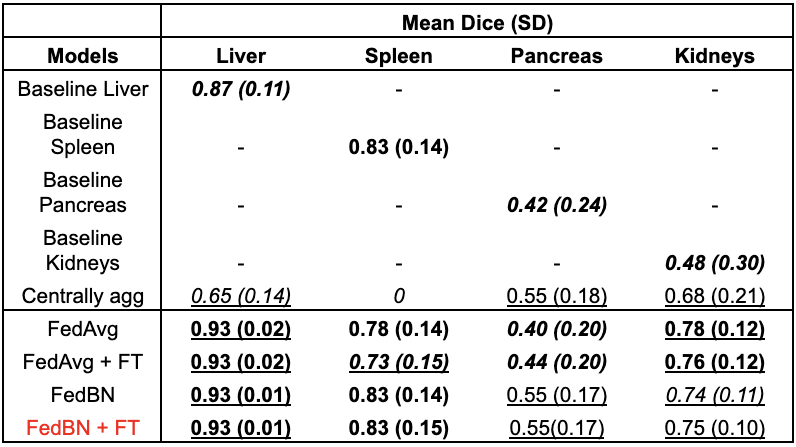}
    \label{tab:btcv_metrics}
\end{table}

\section{Discussion}
The SegViz framework proposed in this work demonstrated excellent performance in aggregating knowledge from heterogeneous datasets with different, incomplete labels. Our approach successfully aggregated knowledge from all nodes with little to no drop in the performance of the global meta-model in terms of the average dice score. The comparable performance between the SegViz segmentations and multiple baseline model segmentation illustrates a preliminary example of constructing a single global multi-task segmentation model with clinical applicability from dispersed datasets with disjoint partial annotations. It is important to note that the FedAvg global model can be extended to contain a multi-head classifier block while this is not true for the FedBN model. 

Image segmentation from heterogeneous datasets with incomplete annotations has many potential benefits. For example, SegViz can potentially reduce labeling time by $1/\eta$ where $\eta$ is the number of distinct labels in the distributed data sets by allowing the transfer of knowledge between each client. This would not only save time but also allow different research groups to potentially benefit from each others’ annotations without explicitly sharing them. 

We believe the success of SegViz is attributed to several inherent advantages in its implementations, such as using a learning rate decay and random affine transformations during training which makes it more robust to non-i.i.d data. Moreover, extending our FL implementations with fine-tuning allows for creating stable, high-performing personalized local models. In the future, we would like to extend our experiments using a modality than is less stable than CT such as MRI. We would also like to investigate the real-world performance of our FL setup where client nodes can join and drop contact with the server at any point in time while maintaining no drop in performance.

\bibliography{bibliography}




% LNCS Homepage, \url{http://www.springer.com/lncs}. Last accessed 4
\end{document}